\documentclass[twocolumn]{article}

\usepackage{amssymb}
\usepackage{booktabs} 
\usepackage{amsmath} 
\usepackage{booktabs} 
\usepackage{caption}  
\usepackage{ragged2e} 
\usepackage{array}    
\usepackage{subcaption} 
\usepackage{graphicx}

\usepackage{cite}

\usepackage{listings}
\begin{document}



\title{Positional Bias in Binary Question Answering: How Uncertainty Shapes Model Preferences}

\author{Tiziano Labruna$^{1}$, Simone Gallo$^{2}$, Giovanni Da San Martino$^{1}$\\
$^1$University of Padova, Italy \quad $^2$CNR-ISTI, Pisa, Italy}

\date{}

\maketitle

\begin{abstract}
Positional bias in binary question answering occurs when a model systematically favors one choice over another based solely on the ordering of presented options. In this study, we quantify and analyze positional bias across five large language models (LLMs) under varying degrees of answer uncertainty. We re-adapted the \textsc{SQuAD-it} dataset by adding an extra incorrect answer option and then created multiple versions with progressively less context and more out-of-context answers, yielding datasets that range from low to high uncertainty. Additionally, we evaluate two naturally higher-uncertainty benchmarks: (1) \textsc{WebGPT} question pairs with unequal human-assigned quality scores, and (2) \textsc{Winning Arguments}, where models predict the more persuasive argument in Reddit’s r/ChangeMyView exchanges. Across each dataset, the order of the “correct” (or higher-quality/persuasive) option is systematically flipped (first placed in position 1, then in position 2) to compute both Preference Fairness (PF) and Position Consistency (PC). We observe that positional bias is nearly absent under low‐uncertainty conditions, but grows exponentially when it becomes doubtful to decide which option is correct.

\end{abstract}



\section{Introduction}
Large language models (LLMs) have demonstrated impressive capabilities in a wide range of natural language understanding and generation tasks, including open‐domain question answering (QA), summarization, and dialogue. However, their behaviors are not always aligned with human expectations of fairness and robustness. One pervasive phenomenon is \emph{positional bias}: the tendency of a model to prefer one answer over another based purely on the position in which each option is presented, rather than on semantic merit. Positional bias can lead to systematic errors when models are asked to choose between two or more alternatives and may undermine trust in their outputs when reliability is critical (e.g., in legal or medical contexts, or when aggregating candidate responses).

Prior work has documented position bias in classification and question-answering tasks \cite{zheng2023judging, wang2024eliminating, dominguez2024questioning, zhu2023judgelm, li2024anchored}. Yet, a systematic study of how positional bias scales with \emph{answer uncertainty} (the degree to which a model can confidently distinguish between options) remains lacking. Intuitively, under low‐uncertainty conditions (e.g., when one answer is clearly correct and context is provided), a well‐trained model should consistently select the correct answer regardless of its placement. As uncertainty rises, through removal of context or through creating two equally plausible (or equally out‐of‐context) options, models may increasingly resort to spurious heuristics, including simply favoring the first or second listed choice. Understanding this phenomenon is critical for: (1) diagnosing model weaknesses, (2) developing evaluation benchmarks that detect fragile behaviors, and (3) designing interventions that mitigate positional bias in downstream applications.

In this work, we conduct a comprehensive investigation of positional bias under varying degrees of uncertainty and across five state-of-the-art LLMs: Llama-3.1-8B, Gemma-3-12B (quantized), Gemini-1.5, Gemini-2, and Phi4-14B (quantized). 

We re‐adapted \textsc{SQuAD-it} \cite{10.1007/978-3-030-03840-3_29} by generating binary question–answer pairs to create a series of benchmarks with controlled uncertainty. The result is an expanded dataset, which we call SQuAD-it-2. We decided to produce this dataset in Italian, as it is a language that remains largely underrepresented in studies on positional bias and answer ordering, allowing us to test models in a setting where linguistic priors are less well-anchored.
First, we include the context and a plausible but incorrect distractor (low uncertainty). Next, we remove the context so that the model must choose between the correct answer and the distractor without supporting evidence (medium uncertainty). Finally, we present two out‐of‐context distractors in place of the correct answer (high uncertainty). We publicly release all generated versions.\footnote{https://github.com/tLabruna/SQuAD-it-2}

Additionally, we identified two datasets that involve subjective judgments or nuanced quality comparisons. The first is \textsc{WebGPT}~\cite{nakano2021webgpt}, which provides human‐rated preferences between pairs of model‐generated answers to the same question. The second is \textsc{Winning Arguments}~\cite{tan+etal:16a}, featuring pairs of Reddit r/ChangeMyView responses to a single post, where only one reply earned a “delta” for being deemed more persuasive.

Across these datasets, we measure positional bias using two complementary metrics (Preference Fairness and Position Consistency) that capture whether and how often a model’s decision changes when the order of the candidate answers is swapped. Through the experiments we uncover a clear pattern: positional bias is negligible when uncertainty is low but grows exponentially as uncertainty increases. Moreover, we find that this effect is especially pronounced in tasks requiring subjective judgment, where models frequently default to order‐based heuristics in the absence of unambiguous signals.


The remainder of the paper is organized as follows: Section \ref{sec:related} reviews related work on position bias and fairness in NLP. Section \ref{sec:method} details our dataset construction, experimental protocols, and bias metrics. Section \ref{sec:results} presents quantitative results and discusses the significance of the outcomes. Finally, Section \ref{sec:conclusion} concludes and outlines future directions.

\begin{figure*}[ht]
    \centering
    \includegraphics[height=0.3\textheight]{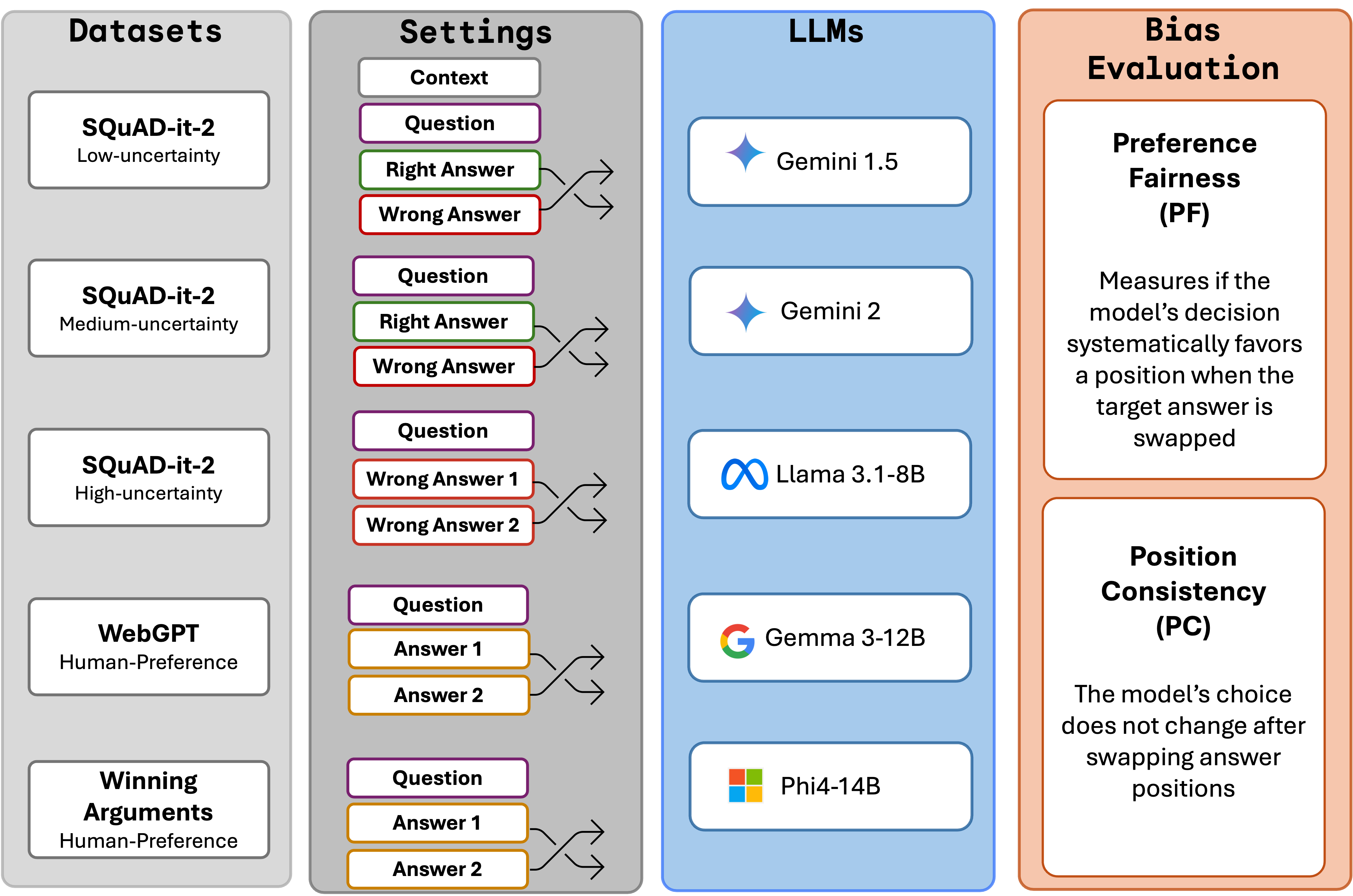}
    \caption{Overview of the five datasets used in the study, including the settings for each dataset: for the SQuAD datasets, each includes two answers to a question, one correct and one incorrect; for WebGPT and Winning Arguments, each dataset consists of two possible messages with one annotated as higher quality or more persuasive. The figure also shows the five LLMs evaluated, along with the two positional bias metrics used: Preference Fairness (PF) and Position Consistency (PC).}
    \label{fig:fullpage}
\end{figure*}

\section{Related Work}
\label{sec:related}
The growing adoption of Large Language Models (LLMs) in both generation and evaluation tasks has brought increased scrutiny to their fairness, especially in contexts involving binary or pairwise decisions. A prominent concern is positional bias—a systematic preference for one response over another based solely on its position in the prompt, irrespective of content. Our work builds on and differentiates itself from a body of literature that has examined this phenomenon under various evaluation and reasoning paradigms.

The study by Shi et al. \cite{abs-2406-07791} offers the most comprehensive exploration of positional bias in LLM-based pairwise evaluation. They introduce three core metrics: Positional Fairness (PF), Positional Consistency (PC), and Repetitional Consistency (RC), to systematically assess how the order of candidate responses affects judgement outcomes. Notably, they find that while most models exhibit high repetitional consistency—i.e., deterministic outputs across repeated trials—positional fairness and consistency vary widely across tasks and models. Their findings demonstrate that positional bias becomes especially pronounced when comparing responses of near-equal quality, an observation that directly informs our own approach of varying answer uncertainty to modulate the ambiguity of binary choices.

While Shi et al. focus primarily on models acting as evaluators, Wang et al. \cite{WangLCCZLCKLLS24} provide compelling evidence of position-sensitive scoring even in ostensibly objective comparisons. They show that GPT-4 tends to favour the first answer while GPT-3.5 leans toward the second, irrespective of prompt instruction, as also highlighted by similar studies \cite{casola2023testing}. Their proposed mitigation strategies—including Balanced Position Calibration (BPC) and Multiple Evidence Calibration (MEC)—highlight the importance of structural prompt design in mitigating these biases. Our study similarly adopts systematic answer reordering, but unlike Wang et al., we extend the analysis to task formats beyond pairwise model evaluation, such as QA under uncertainty.

Other work, such as \cite{Zheng0M0H24}, shifts the lens toward multi-option multiple choice settings. The authors distinguish between token bias—a preference for specific answer IDs like “A” or “B”—and position bias—a preference for answers based on ordinal position. Their central claim is that token bias, not positional bias, is the primary cause of inconsistencies in MCQ tasks, and they propose PriDe, a debiasing method based on prior estimation. While they conclude that positional bias is secondary and often overestimated, our findings suggest that under heightened uncertainty, position bias becomes marked, particularly when correct answers are ambiguous and out-of-context.

The PORTIA framework proposed by another recent study \cite{LiW0W0G024} presents an architectural solution to reduce positional dependency by restructuring the input through segmental alignment. Although PORTIA is designed for evaluator settings, its contribution lies in demonstrating that careful content interleaving can dampen reliance on positional heuristics. While our methodology does not employ PORTIA-like restructuring, it shares a core intuition: positional effects intensify when content cues are weak or ill-formed, a condition we explicitly engineer through dataset manipulation.

The CALM framework \cite{YeWHCZMGG0CC025} offers a general-purpose protocol for quantifying a wide range of biases in LLM-as-a-judge settings. Its automated perturbation method—swapping candidate positions to detect volatility in outcomes—serves as a direct methodological precedent for the Position Consistency metric. Moreover, CALM’s observation that positional bias scales with the number of response options aligns with our finding that bias intensifies when answer certainty decreases.

In contrast to all aforementioned works, our study offers a novel synthesis of two research trajectories: binary positional evaluation under uncertainty and large-scale QA-based benchmarking. By systematically controlling for answer ambiguity across datasets derived from SQuAD-it, WebGPT, and Reddit's r/ChangeMyView (Winning Arguments dataset), we demonstrate that positional bias is not merely an artefact of model prompt formatting or answer labelling conventions. Rather, it reflects a deeper tendency of LLMs to resolve ambiguity through positional priors—a phenomenon that expands the scope of prior observations made in evaluation-only contexts. Furthermore our work empirically substantiates the claim that positional bias is conditional—not fixed—and emerges as a second-order inference strategy when primary cues are degraded.

In sum, our contribution lies in bridging the gap between diagnostic evaluator studies and answer-generation tasks, showing that positional bias is neither an isolated nor a negligible phenomenon, but one that is sensitive to context, task framing, and content quality. This dual framing broadens the understanding of bias in LLMs and calls for future work on uncertainty-aware prompt and dataset design.

\section{Methodology}
\label{sec:method}

In this section, we describe the construction of our positional bias benchmarks, the experimental protocol for prompting and evaluation, the set of language models under investigation, and the metrics used to quantify positional bias.

\subsection{Datasets}
\label{subsec:datasets}

To systematically investigate positional bias under varying levels of uncertainty, we constructed a new benchmark suite, \textsc{SQuAD-it-2}, derived from the Italian \textsc{SQuAD-it} dataset \cite{10.1007/978-3-030-03840-3_29} and spanning three uncertainty conditions: Low, Medium, and High. In addition, we employed two existing datasets—\textsc{WebGPT} and \textsc{Winning Arguments}—which capture human preference in more subjective decision-making contexts.

Each dataset is structured around binary-choice instances, represented either as quadruples $(C, Q, A_1, A_2)$ or triples $(Q, A_1, A_2)$, where $C$ is an optional context, $Q$ is a question or prompt, and $(A_1, A_2)$ are candidate answers. One answer is designated as the \emph{preferred} choice, while the other serves as a \emph{distractor}.

\paragraph{\textbf{SQuAD-it-2 Low Uncertainty.}}  
This setting builds upon the \textsc{SQuAD-it} dataset \cite{10.1007/978-3-030-03840-3_29}, a semi-automatic Italian translation of the original English \textsc{SQuAD} dataset \cite{rajpurkar2016squad}. Each sample in \textsc{SQuAD-it} is structured as a triple $(Q, C, A_{\text{corr}})$, where $Q$ is a question, $C$ is a supporting context passage, and $A_{\text{corr}}$ is the correct answer, which is always explicitly contained in the context.

However, for our study on positional bias in binary-choice settings, we needed pairs of answer candidates: one correct and one incorrect. To construct these, we used Gemini-2 to generate a plausible but incorrect answer ($A_{\text{plaus}}$) for each sample in \textsc{SQuAD-it}. Specifically, we prompted Gemini-2 with the context $C$, the question $Q$, and the correct answer $A_{\text{corr}}$, instructing it to generate an alternative answer that is plausible—meaning it could conceivably be a correct answer based on the question, but is in fact incorrect. The exact prompt used is included in Appendix~\ref{appendix:gemini_prompt}.

This resulted in a dataset where each instance takes the form $(C, Q, A_{\text{corr}}, A_{\text{plaus}})$. The presence of the context $C$ provides strong evidence in favor of the correct answer, minimizing ambiguity and uncertainty in the model's decision. This version is intended to simulate the lowest level of uncertainty, where one answer is clearly supported by the context and the other, while plausible, is not. While we generated and publicly released \textsc{SQuAD-it-2} for both training and test splits, we consider only the test set, which includes $7{,}609$ samples, for the experiments of this paper.

\paragraph{\textbf{SQuAD-it-2 Medium Uncertainty.}}  
In this version, we reuse the same set of samples from the Low Uncertainty setting, including the same plausible incorrect answers generated by Gemini-2. However, to increase the level of uncertainty, we deliberately remove the context $C$ from each sample. This modification results in instances of the form $(Q, A_{\text{corr}}, A_{\text{plaus}})$, where the model is asked to choose between two answers without access to the supporting information.

In the absence of context, the task becomes significantly more challenging. While the correct answer remains correct in an absolute sense, the model cannot rely on evidence from the passage to make its choice. Sometimes, the question can still be answered using world knowledge or intuition; other times, it becomes virtually impossible to determine which answer is correct based solely on the question. As a result, this version introduces a medium level of uncertainty, greater than in the contextualized setting, but not entirely arbitrary, since one answer is still grounded in the original question. The dataset comprises 7,609 samples from the test split.

\paragraph{\textbf{SQuAD-it-2 High Uncertainty.}}  
This version represents the maximum level of uncertainty, simulating a scenario in which the model must choose between two equally ungrounded options. Here, we prompt Gemini-2 to generate two completely out-of-context (ooc) answers for each question $Q$. The prompt (included in Appendix \ref{appendix:gemini_prompt}) provides the question, the context and the correct answer, instructing Gemini-2 to generate two answers that are non-plausible, that is, they should not reasonably answer the question and should bear no clear relation to the topic.

The resulting instances are structured as $(Q, A^{(1)}_{\text{ooc}}, A^{(2)}_{\text{ooc}})$, where both answers are distractors. Since neither candidate is appropriate or grounded in the question, there is no clear basis for choosing one over the other. In this setting, the model’s decision is expected to approximate random guessing, and the task itself loses semantic validity. Nonetheless, we include this version to simulate conditions of extreme ambiguity and explore how models behave when confronted with entirely unsupported, content-free binary choices. This allows us to probe the outer limits of positional bias, where no rational basis for preference exists. Also this version includes 7,609 samples from the test split.

Overall, the three \textsc{SQuAD-it-2} variants form a controlled uncertainty spectrum, from minimal ambiguity in the Low Uncertainty setting to total ambiguity in the High Uncertainty setting, enabling us to systematically study how large language models respond to answer ordering under varying epistemic conditions.

\paragraph{\textbf{WebGPT.}} The WebGPT dataset \cite{nakano2021webgpt} was introduced to support research in aligning long-form question answering systems with human preferences. It consists of 19,578 comparisons between pairs of answers to the same open-ended question, each annotated with human preference scores. These answers were originally generated by a GPT-3 model fine-tuned via imitation learning and further optimized using reinforcement learning from human feedback (RLHF). Each comparison includes metadata such as the browsing quotes used to compose the answers and the associated preference scores, which range from $-1$ to $1$ and indicate which answer is preferred by annotators.

For our work, we extracted a subset of this dataset focusing on clear preference signals. Specifically, we selected only those examples in which the two answers received different human scores ($s^{(1)} \ne s^{(2)}$), ensuring a clear distinction between a preferred answer and a less preferred (distractor) one. This yielded to a total of 14,346 samples for our experiments. From each of these selected examples, we constructed input triples $(Q, A_{\text{pref}}, A_{\text{dist}})$, where $Q$ is the original question, $A_{\text{pref}}$ is the answer with the higher human score, and $A_{\text{dist}}$ is the lower-rated alternative. To standardize the task, we reformulated the original human instruction, used during annotation to guide raters in evaluating answer quality, as a prompt question asking the model to choose the better answer.

\paragraph{\textbf{Winning Arguments.}} This dataset \cite{tan+etal:16a} is derived from the \texttt{r/ChangeMyView} subreddit, where users post their opinions and invite others to persuade them to change their views. In this setting, the original poster (OP) can award a “delta” ($\Delta$) to a reply that successfully changed their mind. The dataset contains conversation threads enriched with metadata indicating which replies received a delta, making it a valuable resource for studying persuasion and argument quality.

To construct comparison pairs, the original dataset creators used a controlled pairing strategy: each delta-awarded reply (i.e., persuasive) was matched with the most topically similar reply in the same thread that did not receive a delta (i.e., less persuasive), based on Jaccard similarity. This yields pairs of messages that are highly comparable in content but differ in perceived persuasiveness, allowing fine-grained analysis of what makes one argument more compelling than another. As with WebGPT, this dataset centers on subjective human preferences, making the task inherently uncertain and nuanced.

For our experiments, we used only the test set provided with the dataset, consisting of 807 pairs. Each instance was structured as a triple $(P, M_{\text{pref}}, M_{\text{dist}})$, where $P$ is the original post, $M_{\text{pref}}$ is the reply that received the delta, and $M_{\text{dist}}$ is the similar, non-awarded reply. This dataset adds a valuable dimension to our evaluation by focusing on real-world argumentative discourse and subjective judgments of persuasive effectiveness.

\subsection{Experimental Protocol}
\label{sec:protocol}

We adopt a two‐pass prompting strategy to evaluate positional bias across the five datasets introduced in Section~\ref{subsec:datasets}. The Low and Medium Uncertainty versions of \textsc{SQuAD-it-2} are derived from the same underlying dataset; the difference lies in whether the context is provided: it is included in the Low Uncertainty setting and omitted in the Medium one. All other datasets are evaluated without any context.

Each instance consists of a prompt \(X\), a preferred answer \(A_{\text{pref}}\), and a distractor \(A_{\text{dist}}\). For every evaluation condition, we proceed as follows:

\begin{enumerate}
    \item \textbf{Pass 1 (Original Order).} We construct \textit{Prompt\textsubscript{1}}, placing \(A_{\text{pref}}\) as \emph{Option 1} and \(A_{\text{dist}}\) as \emph{Option 2}, alongside the question and, where applicable, the context. The prompt is submitted to the target model, and its response is recorded as \(C^{(1)}\).

    \item \textbf{Pass 2 (Swapped Order).} We construct \textit{Prompt\textsubscript{2}} by inverting the order of the two answers. The instructional text and context (if any) are kept identical. The model’s response is recorded as \(C^{(2)}\).
\end{enumerate}

Prompt phrasing is tailored to the semantics of each dataset and is reported in Appendix~\ref{app:prompts}. In all cases, the prompts in Pass 1 and Pass 2 are structurally identical except for the position of the two candidate answers. The model’s raw selections \(C^{(1)}\) and \(C^{(2)}\) are logged without transformation and later used in the analysis of positional bias.

\subsection{Models Evaluated}
We benchmark five state‐of‐the‐art large language models (LLMs), selected to cover a spectrum of architectures, parameter scales, and deployment configurations. All models are developed by leading organisations in the field of foundation model research, including both open-weight and proprietary providers.

\begin{itemize}
    \item \textbf{LLaMA-3.1–8B}: An 8-billion-parameter open-weight model \cite{grattafiori2024llama} following the LLaMA architecture, fine-tuned for Italian, and released in late 2024. Its compact size makes it well-suited for downstream use in resource-constrained scenarios.

    \item \textbf{Gemma-3–12B (quantized)}: A 12-billion-parameter open-weight multilingual model, \cite{team2025gemma} quantized to 4-bit precision (Q4\_K\_M) retrieved via the Ollama model hub. This quantised variant is employed for efficiency under computational constraints.

    \item \textbf{Gemini 1.5}: A proprietary multilingual model \cite{team2024gemini} from Google DeepMind, specifically tailored for QA tasks.

    \item \textbf{Gemini 2}: The successor to Gemini 1.5, featuring architectural improvements and retraining on updated corpora.

    \item \textbf{Phi-4–14B (quantized)}: A 14-billion-parameter open-weight multilingual model, \cite{abouelenin2025phi} quantized to 4-bit precision (Q4\_K\_M) retrieved via the Ollama model hub. Like Gemma-3, this model is used in its quantized form to enable evaluation under limited computational resources.
\end{itemize}

Quantized models are adopted primarily due to hardware and latency constraints. To ensure validity, we conducted a preliminary test comparing the quantized and full-precision variants of each model on a 100-instance subset of the \textsc{Winning Arguments} dataset. The results showed almost identical accuracy across both versions, suggesting that quantization does not substantially affect model preference or correctness in our evaluation setting.

\subsection{Bias Metrics}

To quantify positional bias in model preferences, we adopt two significant metrics: \emph{Preference Fairness} (PF), introduced by Shi \emph{et al.}~\cite{abs-2406-07791}, and \emph{Position Consistency} (PC), a widely adopted measure in the positional bias literature and also discussed in their work.

We do not consider \emph{Repetitional Consistency} (RC) (also introduced by Shi \emph{et al.}), which measures model stability across repeated identical queries, as we believe it is not sufficiently related to positional bias and not computable under our two-pass evaluation protocol.

\subsubsection{Preference Fairness (PF)}

PF quantifies \textit{directional positional bias}: the extent to which a model favors one answer position (first or second) independently of content. However, in our setting, we focus on the \textit{magnitude} of this bias, regardless of whether it leans toward the first or second option. To this end, we report the absolute value of the PF score, so that it ranges from 0 (no bias) to 1 (maximal bias).

Formally, we compute a raw PF score (\texttt{PF\textsubscript{raw}}) following Shi \emph{et al.}~\cite{abs-2406-07791}:

$$
\texttt{PF\textsubscript{raw}} = (\texttt{rcn} \times \texttt{irr}) - (\texttt{pcn} \times \texttt{ipr}),
$$

where:
\begin{itemize}
\item \texttt{pcn} is the normalized count of times the model prefers the first (primacy) position.
\item \texttt{rcn} is the normalized count of times the model prefers the second (recency) position.
\item \texttt{ipr} and \texttt{irr} are the fractions of instances where the preferred answer was placed in the first and second position, respectively.
\end{itemize}

To ensure comparability across datasets and evaluation setups, the raw score is normalized using its theoretical minimum and maximum values:

$$
\text{PF} = \left( \frac{\texttt{PF\textsubscript{raw}} - S^{-}_{\min}}{S^{+}_{\max} - S^{-}_{\min}} \right) \times 2 - 1,
$$

where $S^{-}_{\min}$ and $S^{+}_{\max}$ are the minimum and maximum achievable values of \texttt{PF\textsubscript{raw}}, respectively, under the given conditions. This normalization centers the scale around zero and bounds it between $-1$ and $1$.

Finally, we report the absolute value of the resulting PF score:

$$
|\text{PF}| \in [0, 1],
$$

so that:
\begin{itemize}
\item $|\text{PF}| = 0$ indicates no positional bias (preference is content-based and consistent).
\item $|\text{PF}| = 1$ indicates maximum positional bias (model always favors one position regardless of content).
\item Intermediate values reflect increasing degrees of positional influence on preference.
\end{itemize}

\subsubsection{Position Consistency (PC)}

PC assesses \emph{stability} rather than directionality: it measures how often the model selects the same answer before and after the answer order is swapped. Formally:

\[
    \text{PC} \;=\; \frac{1}{N} \sum_{i=1}^{N} \mathbb{I}\bigl( C^{(1)} = C^{(2)} \bigr),
\]

where $C^{(j)} \in \{\text{A}, \text{B}\}$ is the option chosen by the model at pass $j$, and $\mathbb{I}(\cdot)$ is the indicator function.

\begin{itemize}
    \item A value of $\text{PC} = 1$ indicates full positional robustness: the model's choice is unaffected by option order.
    \item Lower values imply that the model’s preference changes depending on which position the answers are presented in.
\end{itemize}

\vspace{1mm}
PF and PC capture orthogonal phenomena: PF indicates \emph{directional preference bias}, while PC reflects \emph{robustness to positional perturbation}. We report both metrics across all model–dataset pairs.

\section{Results and Discussion}
\label{sec:results}

\begin{table*}[]
\centering
\caption{Model performance on binary QA tasks, comparing accuracy when the correct answer is presented first versus second, along with the number of invalid responses for each experiment (i.e., when the model did not produce the expected output format).}
\label{tab:model_performance}

\begin{tabular}{@{}l l r r r r@{}} 
\toprule
\textbf{Dataset} & \textbf{Model} & \multicolumn{2}{c}{\textbf{Correct-first}} & \multicolumn{2}{c}{\textbf{Wrong-first}} \\
\cmidrule(lr){3-4} \cmidrule(lr){5-6}
& & \textbf{Accuracy} & {\textbf{\# Invalid}} & \textbf{Accuracy} & {\textbf{\# Invalid}} \\
\midrule
SQuAD-it-2 & Llama-3.1-8B & 0.940 & 22 & 0.846 & 5 \\
Low Uncertainty & Gemma-3-12B-Q & 0.918 & 0 & 0.907 & 0 \\ 
& Gemini-1.5 & 0.930 & 37 & 0.909 & 10 \\
& Gemini-2 & 0.930 & 17 & 0.913 & 26 \\
& Phi4-14B-Q & 0.923 & 26 & 0.912 & 21 \\ 
\midrule
SQuAD-it-2 & Llama-3.1-8B & 0.662 & 507 & 0.288 & 2026 \\
Medium Uncertainty & Gemma-3-12B-Q & 0.695 & 0 & 0.662 & 0 \\ 
& Gemini-1.5 & 0.765 & 112 & 0.612 & 22 \\
& Gemini-2 & 0.693 & 137 & 0.762 & 132 \\
& Phi4-14B-Q & 0.637 & 209 & 0.761 & 184 \\ 
\midrule
SQuAD-it-2 & Llama-3.1-8B & 0.648 & 1897 & 0.108 & 1849 \\
High Uncertainty & Gemma-3-12B-Q & 0.411 & 0 & 0.590 & 16 \\ 
& Gemini-1.5 & 0.616 & 908 & 0.262 & 940 \\
& Gemini-2 & 0.256 & 1727 & 0.522 & 1701 \\
& Phi4-14B-Q & 0.705 & 420 & 0.288 & 1448 \\ 
\midrule
WebGPT & Llama-3.1-8B & 0.837 & 20 & 0.372 & 17 \\
& Gemma-3-12B-Q & 0.736 & 2 & 0.563 & 0 \\ 
& Gemini-1.5 & 0.791 & 13 & 0.490 & 6 \\
& Gemini-2 & 0.649 & 0 & 0.696 & 2 \\
& Phi4-14B-Q & 0.788 & 23 & 0.505 & 14 \\ 
\midrule
Winning Arguments & Llama-3.1-8B & 0.411 & 11 & 0.758 & 12 \\
& Gemma-3-12B-Q & 0.302 & 0 & 0.823 & 0 \\ 
& Gemini-1.5 & 0.321 & 0 & 0.808 & 0 \\
& Gemini-2 & 0.470 & 0 & 0.766 & 0 \\
& Phi4-14B-Q & 0.178 & 56 & 0.820 & 62 \\ 
\bottomrule
\end{tabular}
\end{table*}

\begin{figure*}[]
  \centering
  \begin{subfigure}[b]{0.49\textwidth}
    \centering
    \includegraphics[width=\textwidth]{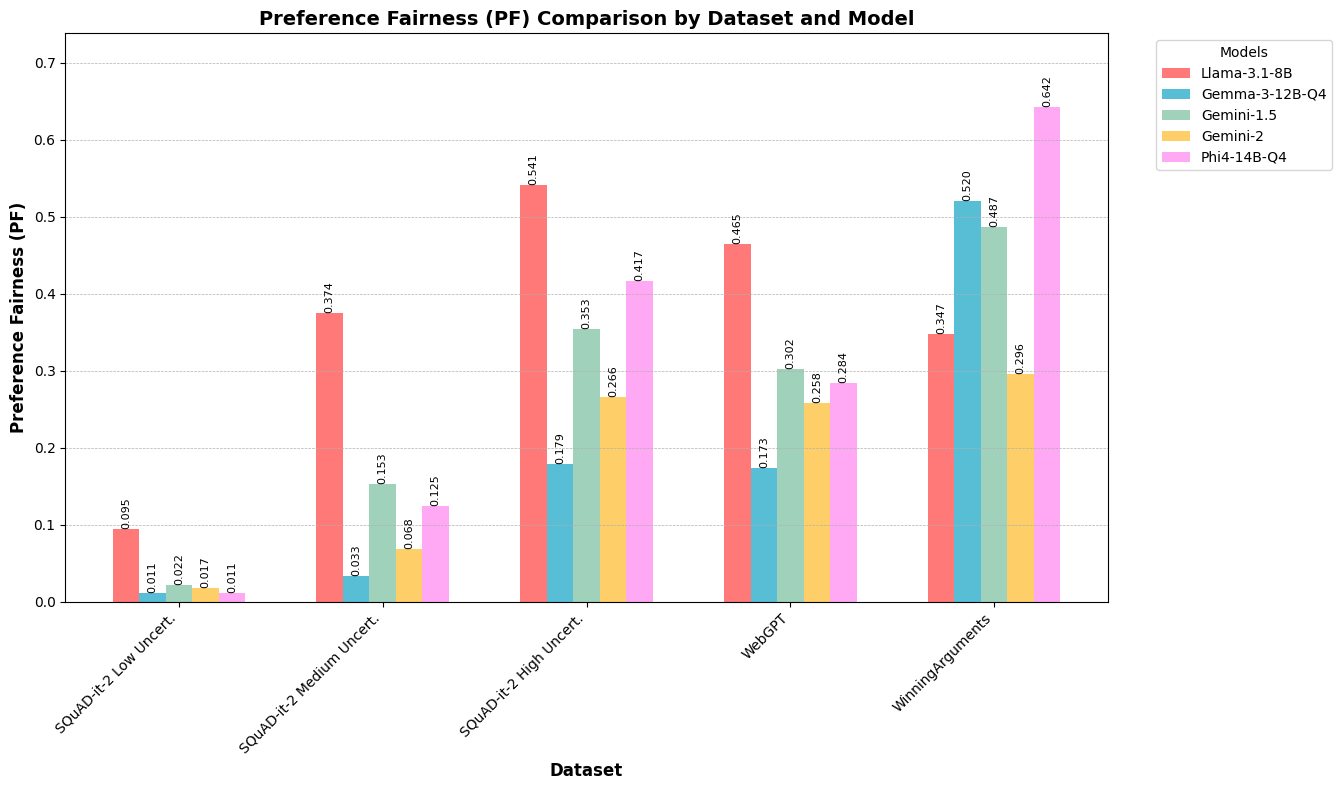}
    \caption{Preference Fairness}
    \label{fig:grafico_pf}
  \end{subfigure}
  \hfill
  \begin{subfigure}[b]{0.49\textwidth}
    \centering
    \includegraphics[width=\textwidth]{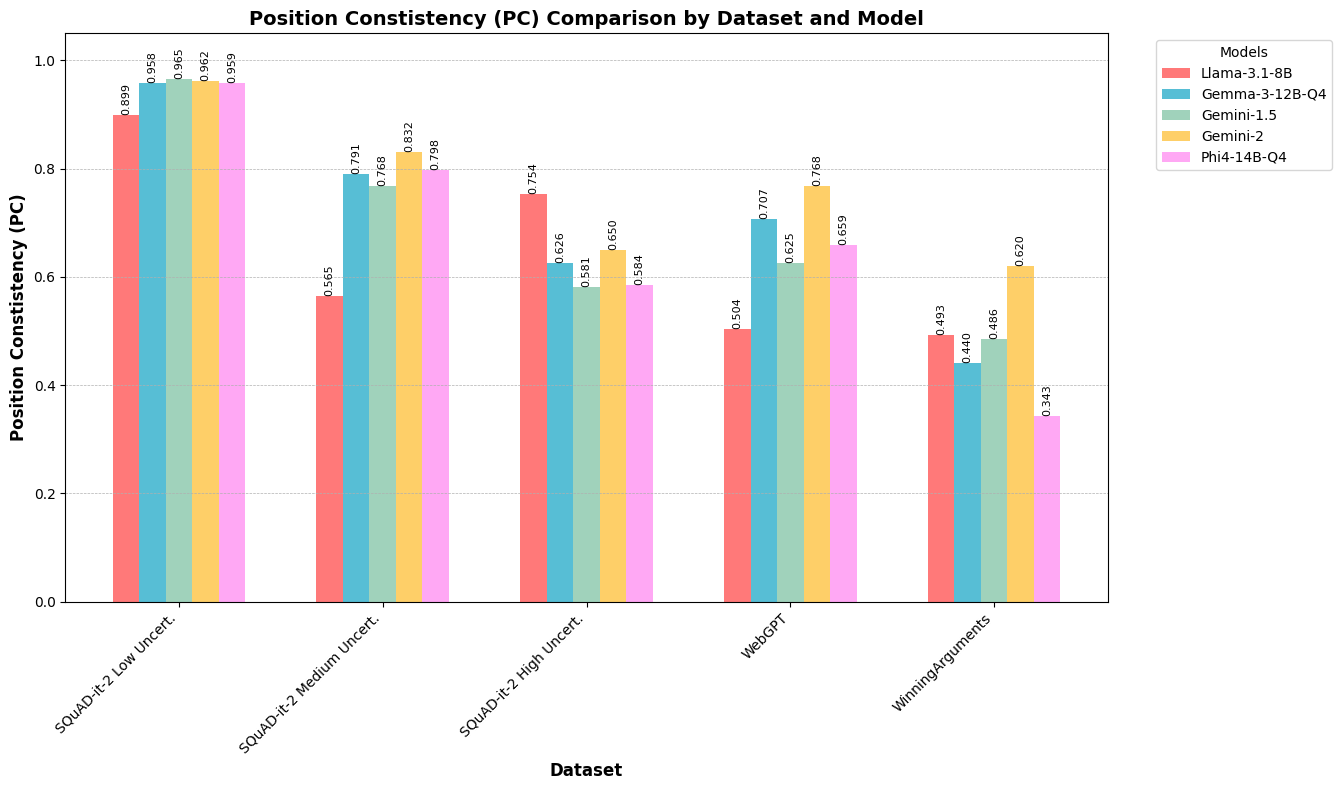}
    \caption{Position Consistency}
    \label{fig:grafico_pc}
  \end{subfigure}
  \caption{Visualization of positional bias across models and datasets, reported by the absolute value of \textbf{Preference Fairness (PF)} (higher absolute values indicate stronger bias) and \textbf{Position Consistency (PC)} (lower values indicate stronger bias).}
  \label{fig:metrics_results}
\end{figure*}

Table~\ref{tab:model_performance} reports the performance of various models on binary QA tasks across datasets with varying levels of uncertainty. Each model is evaluated under two conditions: when the correct answer is presented first and when it is presented second. Additionally, we report the number of \textit{invalid responses}, i.e., outputs not conforming to the expected binary format. Figure~\ref{fig:metrics_results} provides a visualization of the magnitude of positional bias, with bars showing the values of PF (reported in absolute value) and PC for every model and dataset evaluated. In this plot, higher PF values indicate stronger positional bias, while lower PC values correspond to reduced position consistency and thus higher bias. While the figure offers an immediate overview of how bias varies across datasets and models, the accuracy table provides more detailed insights into model behavior, revealing specific patterns such as systematic preference for a given position or consistent shifts in performance depending on answer order.

\paragraph{SQuAD-it-2 Low Uncertainty}

Under low uncertainty, performance is high and relatively stable. All models (except Llama) maintain accuracy above 90\% across both conditions. This indicates that when questions are clear and straightforward, the models perform robustly and are less sensitive to presentation order.

\paragraph{SQuAD-it-2 Medium and High Uncertainty}

As uncertainty increases, performance drops and order effects become more pronounced. In the medium uncertainty setting, accuracy generally decreases across models, and some models (e.g., Gemini-2 and Phi4-14B-Q) actually perform slightly better when the wrong answer is presented first. This may reflect a shift in reliance from positional bias to internal reasoning mechanisms.

In the high uncertainty setting, models diverge sharply. For example, Llama-3.1-8B shows a drastic drop in accuracy when the wrong answer is presented first (from 0.648 to 0.108), indicating a strong sensitivity to order under ambiguous conditions. In contrast, Gemma-3-12B-Q improves when the wrong answer is first (from 0.411 to 0.590), suggesting a different processing dynamic. Invalid responses spike in this setting, especially for Llama and Gemini models, indicating a higher difficulties in producing well-formed answers when the uncertainty is higher.

\paragraph{WebGPT and Winning Arguments}

Real-world datasets present an additional layer of complexity. In the WebGPT task, most models follow the trend observed in synthetic settings: higher accuracy when the correct answer comes first. However, Gemini-2 again deviates from this pattern, performing slightly better in the wrong-first condition.

In the Winning Arguments dataset, which features highly opinionated and subjective content, the reversal is particularly pronounced: all models consistently perform better when the correct answer is presented second. For instance, Gemma-3-12B-Q improves dramatically from 0.302 to 0.823 accuracy in the wrong-first setting. This striking and systematic pattern suggests that models may be influenced not just by answer content but also by presentation dynamics, such as contrastive framing or cumulative reasoning, where the second answer is implicitly treated as a refinement or counterpoint to the first. It is also possible that models trained on internet discussions and dialogues have internalized discourse norms in which stronger or more convincing arguments often follow weaker ones in order to rebut or build upon them. This behavior warrants further investigation, as it may reveal underlying heuristics the models rely on in persuasive or opinionated domains.

\paragraph{General Trends and Considerations}

Across datasets, several consistent patterns emerge, highlighting how model behavior in binary QA tasks is influenced by a complex interplay of input uncertainty, answer ordering, and model architecture. Most models perform better when the correct answer is presented first, particularly under low uncertainty conditions, suggesting a tendency to favor the first option when questions are clear and unambiguous. However, in the Winning Argument dataset, which involves persuasive argumentation, all models systematically perform better when the correct answer is presented second. The magnitude and consistency of this reversal suggest a strong bias toward the second option in subjective or argumentative contexts, possibly influenced by discourse structure or rhetorical patterns in the training data.

As uncertainty increases, the impact of answer ordering becomes more marked across models and datasets. While many models demonstrate robustness under low uncertainty, with small performance differences between correct-first and wrong-first conditions, their behavior becomes significantly more unpredictable and order-sensitive with higher uncertainty. This growing sensitivity is particularly evident in Figure~\ref{fig:metrics_results}: as the input becomes more ambiguous or subjective, such as in the SQuAD-it-2 High Uncertainty and Winning Arguments settings, models increasingly deviate from uniform behavior and show strong biases. This trend suggests that models may resort to positional heuristics or discourse-level patterns under stress, rather than relying on semantic fidelity alone. The rise in invalid outputs in these settings further emphasizes the difficulty some architectures face in maintaining output consistency and format constraints when confronted with harder or less structured tasks.

\section{Conclusion}
\label{sec:conclusion}

In this work, we conducted a systematic investigation of positional bias in large language models using binary-choice prompting. We evaluated five different LLMs across both controlled tasks and real-world datasets, and introduced a novel benchmark, \textbf{\textsc{SQuAD-it-2}}, to study this phenomenon in Italian, an underrepresented language in current LLM evaluation efforts. \textsc{SQuAD-it-2} includes binary QA tasks at three uncertainty levels, enabling fine-grained analysis of how answer ordering interacts with ambiguity.

Our findings reveal a clear trend: as input uncertainty increases, so does positional bias. Under low uncertainty, models exhibit high accuracy and almost identical performance whether the correct answer is presented first or second, indicating minimal or no bias in these conditions. However, as uncertainty rises, due to the removal of contextual cues or the subjective nature of the task, models begin to show strong and often inconsistent positional preferences.

We used two dedicated metrics to quantify these effects: \textit{Preference Fairness} (PF), which captures how much a model favors one position over another, and \textit{Position Consistency} (PC), which reflects how stable model decisions are across different answer orderings. Both metrics show clear deterioration as uncertainty increases, confirming that models rely more heavily on position-based heuristics when semantic cues are weak.

A particularly striking result comes from the \textsc{Winning Arguments} dataset, where all models systematically prefer the second option---even when it is incorrect. This behavior suggests that models may be influenced not only by answer content but also by presentation dynamics, such as contrastive framing or cumulative reasoning, possibly reflecting discourse norms internalized during training, where stronger arguments often follow weaker ones to refine or counter them.

These results expose a fundamental limitation in current LLMs and highlight the need for robust evaluation and debiasing strategies, especially in high-stakes or subjective scenarios. Our release of \textsc{SQuAD-it-2} provides a valuable tool for continued research, offering a scalable and controlled benchmark for assessing positional artifacts, particularly in multilingual contexts.

Future work should explore the mechanisms behind position-based preferences more deeply, with special attention to how models process discourse structure, contrastive reasoning, and pragmatic cues. Better understanding these behaviors will be crucial for developing more interpretable, trustworthy, and bias-resilient models.

\bibliographystyle{plain}
\bibliography{main}

\appendix
\section{Prompts for SQuAD-it-2 Dataset Generation}
\label{appendix:gemini_prompt}

\subsection{Prompt for Low and Medium Uncertainty Settings}
For the \textsc{SQuAD-it-2} Low and Medium Uncertainty variants, we used a single prompt to generate a plausible but incorrect answer. The input to the model includes the original context passage, the question, and the correct answer. The model is explicitly instructed to generate an answer that could reasonably be interpreted as correct (i.e., plausible), while being in fact incorrect. The exact prompt is shown below:

\begin{lstlisting}
Contesto: <CONTEXT>
Domanda: <QUESTION>
Risposta corretta: <CORRECT_ANSWER>

Fornisci una risposta plausibile ma sbagliata alla domanda sopra, basandoti sul contesto. Restituisci solo la risposta, senza spiegazioni o altro.
\end{lstlisting}

This prompt ensures that the incorrect answer remains semantically coherent with the question and context, but does not match the correct answer.

\subsection{Prompting Strategy for High Uncertainty Setting}
For the High Uncertainty setting, we followed a two-step prompting process to construct a pair of out-of-context (OOC) incorrect answers. The correct answer is used during generation but removed from the final dataset to increase ambiguity.

\paragraph{Step 1: First Out-of-Context Answer.}  
The model receives the context, the question, and the correct answer. It is asked to generate an incorrect answer that is completely unrelated to the provided context, ensuring it is not plausible or grounded. The prompt used is:

\begin{lstlisting}
Contesto: <CONTEXT>
Domanda: <QUESTION>
Risposta corretta: <CORRECT_ANSWER>

Fornisci una risposta completamente fuori contesto e sbagliata alla domanda sopra. Assicurati che non sia basata sul contesto fornito. Restituisci solo la risposta, senza spiegazioni o altro.
\end{lstlisting}

\paragraph{Step 2: Second Out-of-Context Answer.}  
The model is then prompted again with the same context, question, and correct answer, along with the previously generated out-of-context wrong answer. This time, it is asked to produce a second, distinct out-of-context answer. The corresponding prompt is:

\begin{lstlisting}
Contesto: <CONTEXT>
Domanda: <QUESTION>
Risposta corretta: <CORRECT_ANSWER>
Risposta errata: <WRONG_ANSWER_1>

Fornisci una risposta completamente fuori contesto e sbagliata alla domanda sopra. Assicurati che non sia basata sul contesto fornito e che sia diversa dalla risposta errata gia' presente. Restituisci solo la risposta, senza spiegazioni o altro.
\end{lstlisting}

\paragraph{Final Construction.}  
Once both out-of-context answers are generated, we discard the original context and the correct answer, retaining only the question and the two OOC distractors. The final dataset entries are structured as:

\[
(Q, A^{(1)}_{\text{ooc}}, A^{(2)}_{\text{ooc}})
\]

This setup simulates maximal uncertainty, as neither of the candidate answers is relevant or correct, forcing the model to rely solely on positional priors or heuristics.

\section{Prompt Templates}
\label{app:prompts}

We report here the prompt templates used in the experiments described in Section~\ref{sec:protocol}. Each dataset required a prompt adapted to its semantic framing and language.

\paragraph{\textsc{SQuAD-it-2} (Low and Medium Uncertainty).}
These prompts are in Italian. When context is present (Low Uncertainty), it is introduced with “\texttt{Contesto:}”. The rest of the prompt follows this structure:
\begin{quote}
\texttt{Domanda: [Q]}\\
\texttt{A) [Risposta 1]}\\
\texttt{B) [Risposta 2]}
\end{quote}
The final instruction depends on the uncertainty level:
\begin{itemize}
    \item \textbf{Low Uncertainty (with context):}\\
    \texttt{Scegli la risposta corretta. Restituisci solo A o B.}
    \item \textbf{Medium Uncertainty (no context):}\\
    \texttt{Scegli la risposta che reputi più corretta. Se credi che nessuna sia corretta, scegli comunque quella che reputi più plausibile. Restituisci solo A o B.}
\end{itemize}

\paragraph{\textsc{WebGPT}.}
The prompt is in English and asks the model to determine which answer is more useful:
\begin{quote}
\texttt{You are given a question and two answers, A and B. Your task is to decide which answer is overall more useful. Read the question and both answers carefully. Compare them based on how well their claims are supported, how relevant they are to the question, how much unsupported or irrelevant content they include, and how coherent and well-written they are. Weigh all these factors and respond with A or B, depending on which answer is better. Do not explain your choice. Output only A or B.}\\

\texttt{Question: [Q]}\\
\texttt{A) [Answer A]}\\
\texttt{B) [Answer B]}\\
\texttt{Answer:}
\end{quote}

\paragraph{\textsc{Winning Arguments}.}
The prompt asks the model to judge persuasiveness:
\begin{quote}
\texttt{You are a Persuasion Detector, your goal is to understand if a message is more or less persuasive than another, meaning that it has more or less potential of changing someone's opinion. You will be prompted with 2 messages and you have to respond with ONLY "Message 1" or "Message 2" based on which message you think is more persuasive.}\\

\texttt{---- Message 1: ----}\\
\texttt{[Message 1]}\\

\texttt{---- Message 2: ----}\\
\texttt{[Message 2]}\\

\texttt{Answer:}
\end{quote}

\end{document}